\pdfoutput=1

\documentclass[11pt]{article}

\usepackage{EMNLP2022}

\usepackage{times}
\usepackage{latexsym}

\usepackage[T1]{fontenc}

\usepackage[utf8]{inputenc}

\usepackage{microtype}

\usepackage{inconsolata}
\usepackage{xcolor}
\usepackage{listings}
\usepackage{xparse}
\NewDocumentCommand{\codeword}{v}{\texttt{\textcolor{black}{#1}}}

\title{The Massively Multilingual Natural Language Understanding 2022 (MMNLU-22) Workshop and Competition}

\author{Chris Hench \\ \texttt{henchc@amazon.com} \\\And Charith Peris \\ \texttt{perisc@amazon.com} \\\AND Jack FitzGerald \\ \texttt{jgmf@amazon.com} \\\And Kay Rottmann \\ \texttt{krrottm@amazon.com}}

\begin{document}
\maketitle
\begin{abstract}
Despite recent progress in Natural Language Understanding (NLU), the creation of multilingual NLU systems remains a challenge. It is common to have NLU systems limited to a subset of languages due to lack of available data. They also often vary widely in performance. We launch a three-phase approach to address the limitations in NLU and help propel NLU technology to new heights. We release a 52 language dataset called the Multilingual Amazon SLU resource package (SLURP) for Slot-filling, Intent classification, and Virtual assistant Evaluation, or MASSIVE, in an effort to address parallel data availability for voice assistants. We organize the Massively Multilingual NLU 2022 Challenge to provide a competitive environment and push the state-of-the art in the transferability of models into other languages. Finally, we host the first Massively Multilingual NLU workshop which brings these components together. The MMNLU workshop seeks to advance the science behind multilingual NLU by providing a platform for the presentation of new research in the field and connecting teams working on this research direction. This paper summarizes the dataset, workshop and the competition and the findings of each phase.

\end{abstract}

\section{Introduction}

According to a 2020 study by Juniper Research \cite{juniper} it is expected that by 2024 there will be over 8 billion virtual assistants worldwide, the majority of which will be on smartphones.
Additionally, over 100 million smart speakers have been sold, and virtual assistants continue to be integrated into new products. These devices have in common that humans interact with them via natural language interfaces.
This development has significantly boosted research to advance natural language understanding. 
However, most natural language understanding work focuses on only a few of the more than 4,000 written languages in the world.
The limitation is driven by the lack of labeled data, the expense associated with human-based quality assurance, model maintenance, update costs, and more.
To overcome these hurdles, further research in the field of multilingual natural language understanding is needed to enable natural language understanding for currently not- or under-served languages.
With \citet{MetaNoLanguageLeftBehind} and related work, we have seen progress in recent years on the expansion of machine translation into the domain of under-served languages both by advancing science as well as creation of corpora in machine translation. However, in areas as NLU modeling for virtual assistants, many of these limitations still remain. 
The vision of this workshop is to address the limitations in NLU and help propel NLU technology into the 50-language, 100-language, and even the 1,000-language regime, both for production systems and for research endeavors, succinctly captured by our slogan, {\it Let’s scale natural language understanding technology to every language on Earth!}. We do this via a three-pronged approach. First, we created and released the Multilingual Amazon SLU resource package (SLURP) for Slot-filling, Intent classification, and Virtual assistant Evaluation, or MASSIVE dataset \citep{massiveDataset}, containing 1 million realistic, parallel, labeled virtual assistant text utterances spanning 51 languages. Second, we hosted the Massively Multilingual NLU (MMNLU) 2022 Challenge, a competition designed to advance massively multilingual NLU modeling. Finally, we organized the first MMNLU workshop to bring together researchers working in the field of NLU. By providing much needed labelled data, motivating multilingual NLU exploration and bringing NLU researchers together to share findings and spark further collaboration, we hope to push the state-of-the art in multilingual natural language understanding technology.

\section{Workshop overview}

The first MMNLU 2022 workshop is co-located with EMNLP 2022 in Abu Dhabi. In our call for papers we asked for submissions relevant to the advancement of the field of multilingual NLP. We were particularly interested in submissions related to the shared tasks part of this workshops competition on the recently published MASSIVE \cite{massiveDataset} dataset (see \ref{sec:massive:dataset}) or other multilingual data-sets. We sought work exploring multilingual representations, augmentation and pre-processing techniques, and more efficient models.

\section{Paper Submissions}
In total we received 12 submissions for the venue, of which 8 were accepted for presentation at the workshop.
The papers being part of the proceedings for this workshop showed a wide variety of approaches to deal with the problem of massive multilingual systems. The investigations ranged from the evaluation and investigation of tokenization across languages, towards large language model or translation model based data augmentation as well as direct use of translation systems as a natural language understanding solution. Also, investigations how to minimize degradation on other languages when trained only on a small set of languages, as well as how to use consistency regularization as a way to boost performance were part of the submissions to this workshop.
Another topic investigated by more than one submission was the investigation of code mixing and other cross lingual effects on natural language understanding performance.
Furthermore we also received a paper describing an investigation of how to design templates when Seq2Seq generation is used as a solution for zero-shot cross-lingual tagging.
Overall we are very grateful for the diverse set of research directions proposed in the submissions to this workshop.

\section{The Massive Multilingual NLU 2022 Challenge}

\subsection{MASSIVE dataset}
\label{sec:massive:dataset}
The MASSIVE dataset was localized across 50 languages from the original English data released in the SLURP NLU dataset \cite{slurpDataset}. Unique ids were preserved to yield a parallel corpus, allowing for various natural language understanding tasks beyond intent and slot recognition, such as machine translation. The dataset comprises 60 intents and 55 slot types across 18 domains. The released dataset was partitioned into 587k training utterances, 104k development utterances, and 152k test utterances, also preserving the split used by the original SLURP dataset. For the MMNLU-22 competition, an additional 153k utterances were held-out for the leaderboard. The held-out utterances were created by professionals manually paraphrasing a random sample of SLURP utterances in English, which were subsequently localized along with the original dataset. This resulted in more challenging utterances for NLU, with 49\% more slots per utterance on average.

\subsection{The Competition} 

The MMNLU 2022 Challenge is designed to advance the state of the art of massively multilingual NLU, in which a single model can understand and parse text inputs from many different languages. The competition is based on the recently published MASSIVE \cite{massiveDataset} dataset (see Section~\ref{sec:massive:dataset}). 

The competition consisted of two tasks; namely (1) the Full Dataset Task (\ref{sec:competition:full}) and (2) the Zero Shot task (\ref{sec:competition:zero}). The competition was geared towards two awards: the top-scoring award for the system with the best performance, which was awarded separately for each task, and the organizer’s choice award, which was awarded after considering overall submissions. The competition ran from July 24, 2022 until September 3, 2022.

\subsubsection{Full Dataset Task}
\label{sec:competition:full}
For the Full Dataset Task, a single model, trained on all languages of MASSIVE and according to the given split and hold out data constraints, was evaluated on all languages of the MASSIVE hidden evaluation set consisting of 3,000 utterances per locale. All encoder-only models were required to have fewer than 350M parameters and all sequence-to-sequence models fewer than 700M parameters, including word embeddings. We permitted any data to be used for training, but they must have been publicly available. If not publicly available, or a third party service such as machine translation was utilized, we requested these data be made public. Only the training split was permitted for training, development and test partitions were not. No use of the text in the evaluation set for training was permitted.


\subsubsection{Zero Shot Task}
\label{sec:competition:zero}
For the Zero Shot Task, a single model, trained on only the English training partition of MASSIVE and according to all other given constraints in the Full Dataset Task, was evaluated on every language except English in the MASSIVE hidden evaluation set.


\subsection{Top-scoring award}
\label{sec:top:award}
The top-scoring award was intended to encourage teams to create models within the constraints outlined for the competition (see Sections~\ref{sec:competition:full} and \ref{sec:competition:zero}) while demonstrating the best performance for a given task. The performance of submissions to the top-scoring award were evaluated based on Exact Match Accuracy (EMA) of the intent and slot labels in the predicted utterances provided by the competitor, when matched against the labels in the ground-truth (i.e., the MASSIVE hidden evaluation set).

The two teams that achieved first place on the leaderboard for the two tasks were announced as winners (see Sections~\ref{teamhitscir} and ~\ref{teamfabt5} for winning teams and descriptions of their approaches).

\subsection{Organizer’s choice award}
\label{sec:org:award}
In addition to simple utility, we also wanted to encourage creative approaches to solving the problem of intent classification and named entity recognition. The organizers’ choice award was based primarily on the assessment of the promise of a given approach, and not purely on its leaderboard performance. The assessment was made by a panel of reviewers that consisted of the Program Committee and Organizer’s of MMNLU. See Section~\ref{teambolleke} for the winning team and a description of their approach.
 
\subsection{Leaderboard}
Our leaderboard\footnote{https://eval.ai/web/challenges/challenge-page/1697/leaderboard} was setup on eval.ai \citep{Yadav2019EvalAITB} a centralized platform that hosts Artificial Intelligence (AI) challenges across the globe with the intention of supporting better benchmarking in AI. eval.ai allows for two types of challenge to be hosted on their servers, namely (1) prediction upload challenges, and (2) code upload challenges. 

The MMNLU competition took the form of a prediction upload challenge where competitors were required to train their own models, score them on a hidden evaluation set and then upload the predictions onto the eval.ai challenge. Once uploaded the predictions were run through an evaluation script. The evaluation script was designed to calculate exact match accuracy (EMA), intent accuracy, slot F1, the EMA of the highest performing language and the EMA of the lowest performing language, against the ground truth of the hidden evaluation set.

\subsection{Submissions}
\label{sec:submissions}

We received seven submissions to the Full Dataset task and eight submissions to the ZeroShot task. A total of 11 research teams participated across the two tasks. A top-scoring award was given to the two teams that led the leaderboard for each task, and the organizer’s choice award was given based on the assessment of the promise of a given approach. We briefly describe the winning submissions in the following sections.

\subsubsection{Team HIT-SCIR} 
\label{teamhitscir}
Team \codeword{HIT-SCIR} won the top-scoring award for the Full Dataset Task (see Sections~\ref{sec:top:award} and~\ref{sec:competition:full}). They used a consistency regularization approach with a hybrid data augmentation strategy that included machine translation and subword sampling\footnote{Subword sampling is to apply the on-the-fly subword sampling algorithm in the unigram language model to generate multiple tokenized subword sequences}. Consistency regularization, applied via symmetric Kullback-Leibler divergence, was used to encourage the predicted distributions for an example and its semantically equivalent augmentation to agree with each other.

For the intent detection task, the original example and the predicted distributions of the augmentation data from both strategies are directly aligned. For the slot filling task, the original example can only be aligned with the predicted distribution from the subword sampling augmentation strategy. Slot consistency is ignored when using augmented data from machine translation.

For the Full Dataset task, they used the mT5-base text-to-text model presented by \citet{massiveDataset}, which contains 580M parameters, including 190M embedding parameters. They used examples with the same id in different languages as machine translation augmentation (see \citealt{massiveDataset} for details on the MASSIVE dataset id).

For the Zero Shot task, they used the XLM-align-base model, which contains 270M parameters, including 190M embedding parameters, and a simple two-layer classification head for both intent detection and slot-filling tasks. They used commercial translation APIs to obtain slot-aligned translations and intent-aligned translations.

The consistency regularization-based method does not introduce any additional parameters into the model.

For a detailed description of their submission, refer Team \codeword{HIT-SCIR}'s work \citep{Zheng2022HIT}.

\subsubsection{Team FabT5}
\label{teamfabt5}
Team \codeword{FabT5} won the top-scoring award for the Zero Shot Task (see Sections~\ref{sec:top:award} and~\ref{sec:competition:zero}). They used ByT5 base \citep{Xue2022ByT5TA}, a text-to-text model that takes an input query and outputs a full interpretation composed of the intent, relevant slot labels, and their corresponding slot values.
ByT5 base has the same number of parameters as mT5 base (582M), but distributed differently. While mT5 base has 66\% of its parameters allocated for the embeddings, ByT5 base has only 0.1\% of its parameters allocated for the same purpose.
They trained the ByT5 model to predict an MTOP-style interpretation from a given query. These predictions were then converted back into the intent and annotated utterance field formats in the MASSIVE dataset \citep{massiveDataset}.

In addition, they found that prepending both query and target interpretation with the language string of the query was slightly helpful.
 
For the zero-shot submission the team trained the model using the English MASSIVE training split. To obtain data in the target languages, they translated the English queries using the Google Translate API and projected the slot annotations from the original English queries to the corresponding translations.
To project the annotations, they used the Translate-and-Fill approach, where an mT5 filler model trained on the English queries was used to project the labels to the translations in a zero-shot fashion. The English train partition was the only one used for zero-shot training. The English validation set was used to select the best checkpoint for inference. No hyperparameter tuning was performed. A fixed learning rate of 0.0001 and a batch size of 128 was used for training.

For a detailed description of their submission, refer Team \codeword{FabT5}'s work \citep{Nicosia2022Eval}.


\subsubsection{Team bolleke} 
\label{teambolleke}
Team \codeword{bolleke} won the organizer's choice award (Section~\ref{sec:org:award}). They repurposed a translation model for the task of intent detection and slot filling. The existing dataset was first expanded by generating new training examples via the use of two fine-tuned GPT-3 models. One variant of GPT-3 (13B) was used to generate utterances conditioned on the intent, and a second variant was used to do the intent and slot filling task. Using the two models added confidence via intent agreement. The team generated 20K (English) examples using this method. 
Next, a translation model was trained on the MASSIVE training partition to translate English annotated utterances into the 50 available languages, resulting in an augmented dataset of an additional 1M examples. 
For both pre-training and fine-tuning, the team used an NLLB-200's (No Language Left Behind) distilled 600M parameter variant \citep{MetaNoLanguageLeftBehind}. The objective of the pre-training step was a cross-lingual translation task, with intent detection and slot filling (for example, they trained the model to translate {\it "wat is het weer in new york"} to {\it "weather\_query|quel est le temps \`a [place\_name : new york]"}). The fine-tuning was done on the augmented dataset for another 50K steps with the objective of predicting the annotated utterance in the same language (for example, they train the model to output {\it "weather\_query|what is the weather in [place\_name : new york]"} given the input {\it "what is the weather in new york"}).

Training was done on DeepSpeed with a batch size of 56 and a max length of 64. The input is always the raw utterance and the output the concatenation of the intent and the annotated utterance. Learning rates of 0.0001 and 0.00005 were used for the pre-training and fine-tuning respectively. Given the size of the dataset and computing resources involved they did not engage in hyper-parameter tuning. 

For a detailed description of their submission, refer Team \codeword{bolleke}'s work \citep{DeBruyn2022Machine}.


\section{Conclusions}
\label{sec:conclusions}
This paper presents an overview of the first MMNLU workshop collocated with EMNLP 2022. The paper submissions and competition entries showed encouraging progress in the field of multilingual NLU. We hope that the findings presented as well as the collaborations initiated at this workshop, drives more progress in the field.

\section*{Acknowledgements}
We would like to thank the program committee and the reviewers for their important contributions to the organization of the first MMNLU workshop.


\bibliographystyle{acl_natbib}
\bibliography{custom}

\end{document}